%% file: paper.tex
\documentclass[runningheads]{llncs}
\usepackage{graphicx}
\usepackage{amsmath,amssymb} 
\usepackage{color}

\input{plots}

\usepackage{hyperref}       
\usepackage{url}            
\usepackage{booktabs}       
\usepackage{amsfonts}       
\usepackage{nicefrac}       
\usepackage{amssymb}
\usepackage{amsmath}
\usepackage{multirow}
\usepackage{enumitem}
\usepackage{array}
\usepackage{mathtools}
\usepackage[normalem]{ulem}
\usepackage{wrapfig}
\usepackage{cancel}

\usepackage[outercaption]{sidecap}
\usepackage{floatrow}
\sidecaptionvpos{figure}{m}

\usepackage{tikz}
\usetikzlibrary{shapes.geometric}
\usepackage{rotating}
\usetikzlibrary{positioning}

\usepackage{xcolor}
\usepackage{xspace}

\DeclareMathOperator{\sign}{sign}

\begin{document}
\pagestyle{headings}
\mainmatter
\def\ECCVSubNumber{1793}  

\title{Learning and aggregating deep local descriptors for instance-level recognition} 

\input{abbrev}

\author{Giorgos Tolias\and
Tomas Jenicek\and
Ond{\v r}ej Chum}

\titlerunning{Learning and aggregating deep local descriptors for instance-level recognition}
\authorrunning{G. Tolias \etal}
\institute{Visual Recognition Group, Faculty of Electrical Engineering\\ Czech Technical University in Prague}

\maketitle

\newcommand{\gio}[1]{{\color{blue}{#1}}}
\newcommand{\och}[1]{{\color{cyan}{#1}}}
\newcommand{\orig}[1]{{\color{gray}{#1}}}
\newcommand{\deleted}[1]{}

\begin{abstract}
We propose an efficient method to learn deep local descriptors for instance-level recognition. 
The training only requires examples of positive and negative image pairs and is performed as metric learning of sum-pooled global image descriptors. At inference, the local descriptors are provided by the activations of internal components of the network. We demonstrate why such an approach learns local descriptors that work well for image similarity estimation with classical efficient match kernel methods. The experimental validation studies the trade-off between performance and memory requirements of the state-of-the-art image search approach based on match kernels. Compared to existing local descriptors, the proposed ones perform better in two instance-level recognition tasks and keep memory requirements lower. We experimentally show that global descriptors are not effective enough at large scale and that local descriptors are essential. We achieve state-of-the-art performance, in some cases even with a backbone network as small as ResNet18.

\end{abstract}

\input{intro}
\input{related}
\input{background}

\input{method}
\input{experiments}

\section{Conclusions}
An architecture for extracting deep local features is designed to be combined with ASMK matching. The proposed method consistently outperforms other methods on a number of standard benchmarks, even if a less powerful backbone network is used.
Through an extensive ablation study, we show that the SoA performance is achieved by the synergy of the proposed local feature detector with ASMK.
We show that methods based on local features outperform global descriptors in large scale problems, and also that the proposed method outperforms other local feature detectors combined with ASMK.
We demonstrate why the proposed architecture, despite being trained with image-level supervision only, is effective in learning image similarity based on local features.

\head{Acknowledgements.} The authors would like to thank Yannis Kalantidis for valuable discussions. This work was supported by MSMT LL1901 ERC-CZ grant. Tomas Jenicek was supported by CTU student grant SGS20/171/OHK3/3T/13.

%
%

{\small
\bibliographystyle{splncs04}
\bibliography{egbib}
}
\end{document}

%% file: plots.tex
\usepackage{tikz}
\usetikzlibrary{arrows,shapes,calc,matrix,fit,backgrounds}
\usepackage{pgfplots}
\usepackage{pgfplotstable}
\usepgfplotslibrary{fillbetween}
\pgfplotsset{compat=1.9}

\usepackage{xstring}

\usepgfplotslibrary{external}

\newcommand{\extdata}[1]{\input{#1}}
\IfBeginWith*{\jobname}{fig/extern/}{\finalcopy}{}

\newcommand{\leg}[1]{\addlegendentry{#1}}

\tikzset{every mark/.append style={solid}}
\pgfplotsset{
	grid=both, width=\columnwidth, try min ticks=5,
	every axis/.append style={font=\scriptsize},
	every axis plot/.append style={thick,mark=none,mark size=1.2,tension=0.18},
	legend cell align=left, legend style={fill opacity=0.8},
}

\pgfplotsset{
	dash/.style={mark=o,dashed,opacity=0.7},
	dott/.style={mark=o,dotted,opacity=0.7},
}

%% file: abbrev.tex
\newcommand{\nn}[1]{\ensuremath{\text{NN}_{#1}}\xspace}

\def\roxf{$\mathcal{R}$Oxford\xspace}
\def\rox{$\mathcal{R}$Oxf\xspace}
\def\ro{$\mathcal{R}$O\xspace}
\def\rpar{$\mathcal{R}$Paris\xspace}
\def\rpa{$\mathcal{R}$Par\xspace}
\def\rp{$\mathcal{R}$P\xspace}
\def\rdis{$\mathcal{R}$1M\xspace}

\newcommand\resnet[3]{\ensuremath{\prescript{#2}{}{\mathtt{R}}{#1}_{\scriptscriptstyle #3}}\xspace}

\newcommand{\supe}[1]{^{\mkern-2mu(#1)}}
\newcommand{\dime}[1]{(#1)}

\def\l1{\ensuremath{\ell_1}\xspace}
\def\l2{\ensuremath{\ell_2}\xspace}

\newenvironment{narrow}[1][1pt]
	{\setlength{\tabcolsep}{#1}}
	{\setlength{\tabcolsep}{6pt}}

\newcommand{\ghw}{{\scriptscriptstyle \mbox{\hspace{-1pt}}h\mbox{\hspace{-.5pt}}o\mbox{\hspace{-.5pt}}w\mbox{\hspace{-1pt}}}}
\newcommand{\hatghw}{{\scriptscriptstyle \mbox{\hspace{-1pt}}\hat{h}\mbox{\hspace{-.5pt}}\hat{o}\mbox{\hspace{-.5pt}}\hat{w}\mbox{\hspace{-1pt}}}}
\newcommand{\ioo}{{\scriptscriptstyle \mbox{\hspace{-1pt}}\hat{h}\mbox{\hspace{-.5pt}}o\mbox{\hspace{-.5pt}}w\mbox{\hspace{-1pt}}}}
\newcommand{\ioi}{{\scriptscriptstyle \mbox{\hspace{-1pt}}\hat{h}\mbox{\hspace{-.5pt}}o\mbox{\hspace{-.5pt}}\hat{w}\mbox{\hspace{-1pt}}}}
\newcommand{\iio}{{\scriptscriptstyle \mbox{\hspace{-1pt}}\hat{h}\mbox{\hspace{-.5pt}}\hat{o}\mbox{\hspace{-.5pt}}w\mbox{\hspace{-1pt}}}}
\newcommand{\oii}{{\scriptscriptstyle \mbox{\hspace{-1pt}}h\mbox{\hspace{-.5pt}}\hat{o}\mbox{\hspace{-.5pt}}\hat{w}\mbox{\hspace{-1pt}}}}

\newcommand{\comment} [1]{{\color{orange} \Comment     #1}} 
\newcommand{\commentout}[1]{}

\newcommand{\alert}[1]{{\color{red}{#1}}}
\newcommand{\head}[1]{{\smallskip\noindent\bf #1}}
\newcommand{\equ}[1]{(\ref{equ:#1})\xspace}

\newcommand{\red}[1]{{\color{red}{#1}}}
\newcommand{\blue}[1]{{\color{blue}{#1}}}
\newcommand{\green}[1]{{\color{green}{#1}}}
\newcommand{\gray}[1]{{\color{gray}{#1}}}


\newcommand{\tran}{^\top}
\newcommand{\mtran}{^{-\top}}
\newcommand{\zcol}{\mathbf{0}}
\newcommand{\zrow}{\zcol\tran}

\newcommand{\ind}{\mathbbm{1}}
\newcommand{\expect}{\mathbb{E}}
\newcommand{\nat}{\mathbb{N}}
\newcommand{\zahl}{\mathbb{Z}}
\newcommand{\real}{\mathbb{R}}
\newcommand{\proj}{\mathbb{P}}
\newcommand{\prob}{\mathbf{Pr}}

\newcommand{\mif}{\textrm{if }}
\newcommand{\other}{\textrm{otherwise}}
\newcommand{\minimize}{\textrm{minimize }}
\newcommand{\maximize}{\textrm{maximize }}
\newcommand{\st}{\textrm{subject to }}

\newcommand{\id}{\operatorname{id}}
\newcommand{\const}{\operatorname{const}}
\newcommand{\sgn}{\operatorname{sgn}}
\newcommand{\var}{\operatorname{Var}}
\newcommand{\mean}{\operatorname{mean}}
\newcommand{\trace}{\operatorname{tr}}
\newcommand{\diag}{\operatorname{diag}}
\newcommand{\vect}{\operatorname{vec}}
\newcommand{\cov}{\operatorname{cov}}

\newcommand{\softmax}{\operatorname{softmax}}
\newcommand{\clip}{\operatorname{clip}}

\newcommand{\defn}{\mathrel{:=}}
\newcommand{\peq}{\mathrel{+\!=}}
\newcommand{\meq}{\mathrel{-\!=}}

\newcommand{\floor}[1]{\left\lfloor{#1}\right\rfloor}
\newcommand{\ceil}[1]{\left\lceil{#1}\right\rceil}
\newcommand{\inner}[1]{\left\langle{#1}\right\rangle}
\newcommand{\norm}[1]{\left\|{#1}\right\|}
\newcommand{\frob}[1]{\norm{#1}_F}
\newcommand{\card}[1]{\left|{#1}\right|\xspace}
\newcommand{\diff}{\mathrm{d}}
\newcommand{\der}[3][]{\frac{d^{#1}#2}{d#3^{#1}}}
\newcommand{\pder}[3][]{\frac{\partial^{#1}{#2}}{\partial{#3^{#1}}}}
\newcommand{\ipder}[3][]{\partial^{#1}{#2}/\partial{#3^{#1}}}
\newcommand{\dder}[3]{\frac{\partial^2{#1}}{\partial{#2}\partial{#3}}}

\newcommand{\wb}[1]{\overline{#1}}
\newcommand{\wt}[1]{\widetilde{#1}}

\def\xssp{\hspace{1pt}}
\def\ssp{\hspace{3pt}}
\def\msp{\hspace{5pt}}
\def\lsp{\hspace{12pt}}

\newcommand{\cA}{\mathcal{A}}
\newcommand{\cB}{\mathcal{B}}
\newcommand{\cC}{\mathcal{C}}
\newcommand{\cD}{\mathcal{D}}
\newcommand{\cE}{\mathcal{E}}
\newcommand{\cF}{\mathcal{F}}
\newcommand{\cG}{\mathcal{G}}
\newcommand{\cH}{\mathcal{H}}
\newcommand{\cI}{\mathcal{I}}
\newcommand{\cJ}{\mathcal{J}}
\newcommand{\cK}{\mathcal{K}}
\newcommand{\cL}{\mathcal{L}}
\newcommand{\cM}{\mathcal{M}}
\newcommand{\cN}{\mathcal{N}}
\newcommand{\cO}{\mathcal{O}}
\newcommand{\cP}{\mathcal{P}}
\newcommand{\cQ}{\mathcal{Q}}
\newcommand{\cR}{\mathcal{R}}
\newcommand{\cS}{\mathcal{S}}
\newcommand{\cT}{\mathcal{T}}
\newcommand{\cU}{\mathcal{U}}
\newcommand{\cV}{\mathcal{V}}
\newcommand{\cW}{\mathcal{W}}
\newcommand{\cX}{\mathcal{X}}
\newcommand{\cY}{\mathcal{Y}}
\newcommand{\cZ}{\mathcal{Z}}

\newcommand{\vA}{\mathbf{A}}
\newcommand{\vB}{\mathbf{B}}
\newcommand{\vC}{\mathbf{C}}
\newcommand{\vD}{\mathbf{D}}
\newcommand{\vE}{\mathbf{E}}
\newcommand{\vF}{\mathbf{F}}
\newcommand{\vG}{\mathbf{G}}
\newcommand{\vH}{\mathbf{H}}
\newcommand{\vI}{\mathbf{I}}
\newcommand{\vJ}{\mathbf{J}}
\newcommand{\vK}{\mathbf{K}}
\newcommand{\vL}{\mathbf{L}}
\newcommand{\vM}{\mathbf{M}}
\newcommand{\vN}{\mathbf{N}}
\newcommand{\vO}{\mathbf{O}}
\newcommand{\vP}{\mathbf{P}}
\newcommand{\vQ}{\mathbf{Q}}
\newcommand{\vR}{\mathbf{R}}
\newcommand{\vS}{\mathbf{S}}
\newcommand{\vT}{\mathbf{T}}
\newcommand{\vU}{\mathbf{U}}
\newcommand{\vV}{\mathbf{V}}
\newcommand{\vW}{\mathbf{W}}
\newcommand{\vX}{\mathbf{X}}
\newcommand{\vY}{\mathbf{Y}}
\newcommand{\vZ}{\mathbf{Z}}

\newcommand{\va}{\mathbf{a}}
\newcommand{\vb}{\mathbf{b}}
\newcommand{\vc}{\mathbf{c}}
\newcommand{\vd}{\mathbf{d}}
\newcommand{\ve}{\mathbf{e}}
\newcommand{\vf}{\mathbf{f}}
\newcommand{\vg}{\mathbf{g}}
\newcommand{\vh}{\mathbf{h}}
\newcommand{\vi}{\mathbf{i}}
\newcommand{\vj}{\mathbf{j}}
\newcommand{\vk}{\mathbf{k}}
\newcommand{\vl}{\mathbf{l}}
\newcommand{\vm}{\mathbf{m}}
\newcommand{\vn}{\mathbf{n}}
\newcommand{\vo}{\mathbf{o}}
\newcommand{\vp}{\mathbf{p}}
\newcommand{\vq}{\mathbf{q}}
\newcommand{\vr}{\mathbf{r}}
\newcommand{\vs}{\mathbf{s}}
\newcommand{\vt}{\mathbf{t}}
\newcommand{\vu}{\mathbf{u}}
\newcommand{\vv}{\mathbf{v}}
\newcommand{\vw}{\mathbf{w}}
\newcommand{\vx}{\mathbf{x}}
\newcommand{\vy}{\mathbf{y}}
\newcommand{\vz}{\mathbf{z}}

\newcommand{\vone}{\mathbf{1}}
\newcommand{\vzero}{\mathbf{0}}

\newcommand{\valpha}{{\boldsymbol{\alpha}}}
\newcommand{\vbeta}{{\boldsymbol{\beta}}}
\newcommand{\vgamma}{{\boldsymbol{\gamma}}}
\newcommand{\vdelta}{{\boldsymbol{\delta}}}
\newcommand{\vepsilon}{{\boldsymbol{\epsilon}}}
\newcommand{\vzeta}{{\boldsymbol{\zeta}}}
\newcommand{\veta}{{\boldsymbol{\eta}}}
\newcommand{\vtheta}{{\boldsymbol{\theta}}}
\newcommand{\viota}{{\boldsymbol{\iota}}}
\newcommand{\vkappa}{{\boldsymbol{\kappa}}}
\newcommand{\vlambda}{{\boldsymbol{\lambda}}}
\newcommand{\vmu}{{\boldsymbol{\mu}}}
\newcommand{\vnu}{{\boldsymbol{\nu}}}
\newcommand{\vxi}{{\boldsymbol{\xi}}}
\newcommand{\vomikron}{{\boldsymbol{\omikron}}}
\newcommand{\vpi}{{\boldsymbol{\pi}}}
\newcommand{\vrho}{{\boldsymbol{\rho}}}
\newcommand{\vsigma}{{\boldsymbol{\sigma}}}
\newcommand{\vtau}{{\boldsymbol{\tau}}}
\newcommand{\vupsilon}{{\boldsymbol{\upsilon}}}
\newcommand{\vphi}{{\boldsymbol{\phi}}}
\newcommand{\vchi}{{\boldsymbol{\chi}}}
\newcommand{\vpsi}{{\boldsymbol{\psi}}}
\newcommand{\vomega}{{\boldsymbol{\omega}}}

\newcommand{\rLambda}{\mathrm{\Lambda}}
\newcommand{\rSigma}{\mathrm{\Sigma}}

\def\onedot{.\xspace}
\def\eg{\emph{e.g}\onedot} \def\Eg{\emph{E.g}\onedot}
\def\ie{\emph{i.e}\onedot} \def\Ie{\emph{I.e}\onedot}
\def\cf{\emph{cf}\onedot} \def\Cf{\emph{C.f}\onedot}
\def\etc{\emph{etc}\onedot}
\def\vs{\emph{vs}\onedot}
\def\wrt{w.r.t\onedot} \def\dof{d.o.f\onedot}
\def\etal{\emph{et al}.}

\newcommand{\std}[1]{\tiny{$\pm$#1}}

\makeatother

%% file: intro.tex
\section{Introduction}
\label{sec:intro}
Instance-level recognition tasks are dealing with a very large number of classes and relatively small intra-class variability. 
Typically, even instance-level classification tasks are cast as instance-level search in combination with nearest neighbor classifiers.
The first instance-level search approach to achieve good performance, \ie Video Google~\cite{sz03}, is based on local features and the Bag-of-Words (BoW) representation.
Representing images as collections of vector-quantized descriptors of local features allows for efficient spatial verification~\cite{pci+07}, which turns out to be a key ingredient in search for small objects.
Follow-up approaches improve the BoW paradigm either with finer quantization and better matching schemes~\cite{jds10,taj16,az14} or with compact global descriptors generated through aggregation~\cite{jpd+12,az13}. Good performance is achieved even without spatial verification.

The advent of deep networks made it easy to generate and train global image descriptors. 
A variety of approaches exist~\cite{gar+17,rtc19,agt+15,mmo+16,yyd15,hb16} that differ in the training data, in the loss function, or in the global pooling operation.
However, the performance of global descriptors deteriorates for very large collections of images.
Noh \etal~\cite{nas+17} are the first to exploit the flexibility of global descriptor training in order to obtain local features and descriptors, called DELF, 
for the task of instance-level recognition. 
DELF descriptors are later shown~\cite{rit+18} to achieve top performance when combined with the state-of-the-art image search approach, \ie the Aggregated Selective Match Kernel (ASMK)~\cite{taj16}. Compared to compact global descriptors, this comes with higher memory requirements and search time cost.
In contrast to other learned local feature detectors~\cite{drp+19} that use keypoint-level supervision and non-maxima suppression, DELF features are not precisely localized and suffer from redundancy since deep network activations are typically spatially correlated.

\begin{figure}[t]
\centering
\includegraphics[width=.48\linewidth]{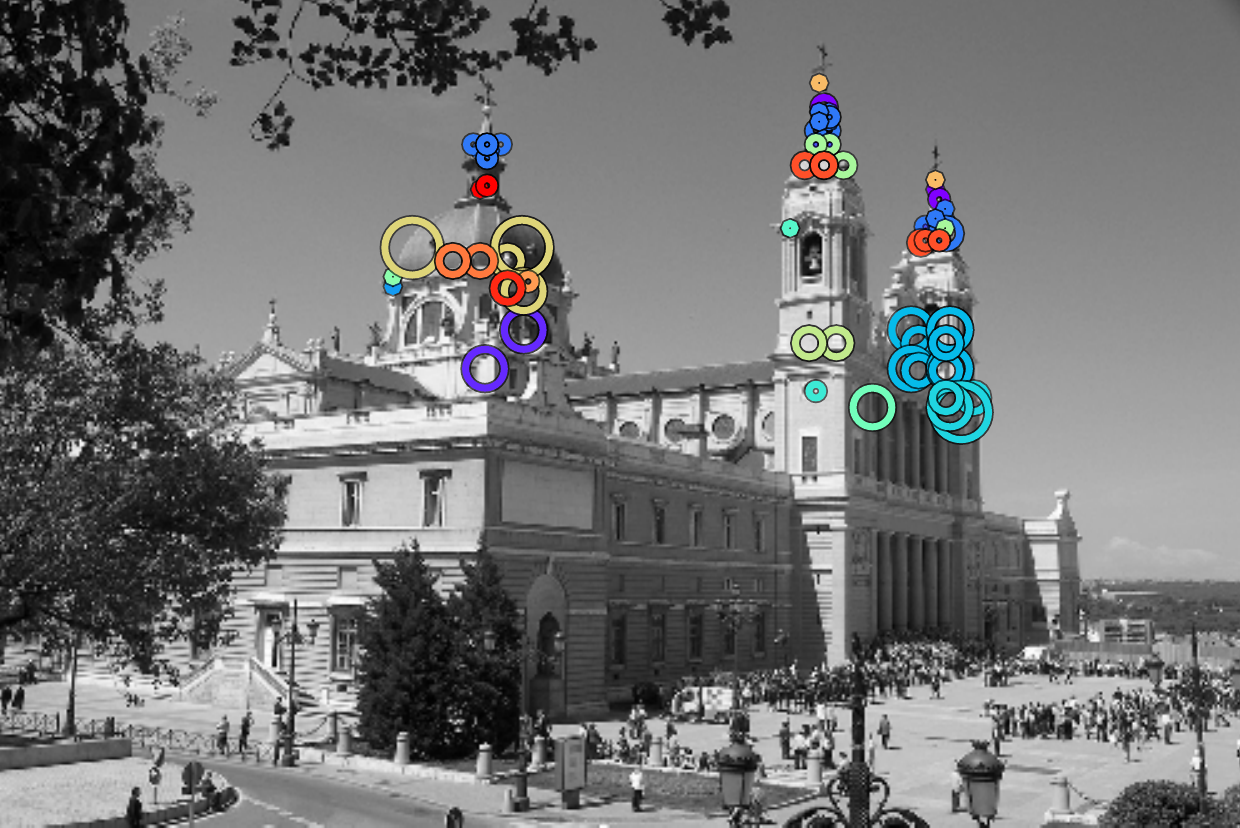}\hfill
\includegraphics[width=.48\linewidth]{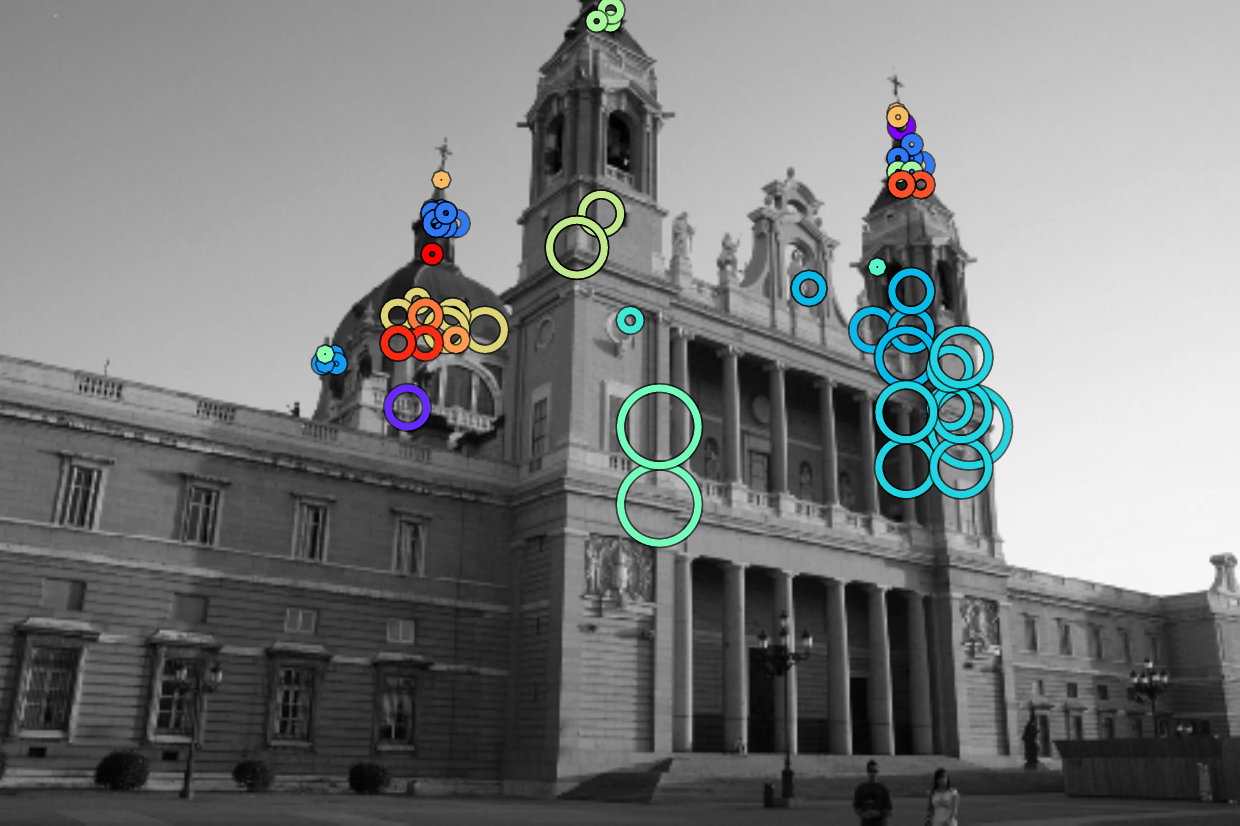}
\caption{Learned local features and descriptors matched with ASMK. 
Features assigned to the same visual word (65k words codebook) are shown in the same color; only top 20 common visual words (out of 94) are included.
Accurate localization is not required since we do not use spatial verification to perform instance-level recognition.
\label{fig:intro}
}
\end{figure}

In this work\footnote{\url{https://github.com/gtolias/how}}, we propose a local feature detector and descriptor based on a deep network. It is trained through metric learning of a global descriptor with image-level annotation.
We design the architecture and the loss function so that the local features and their descriptors are suitable for matching with ASMK, see Figure~\ref{fig:intro}. 
ASMK is known to deliver good performance even without spatial verification, \ie precise feature localization is not crucial, and it deals well with repeated or bursty features.
Therefore, the common drawbacks of existing deep local features for instance-level recognition are overcome.
Unlike classical local features that attempt to offer precise localization to extract reliable descriptors, multiple nearby locations can give rise to a similar descriptor in our training; multiple similar responses are not suppressed, but averaged.

The main contribution of this work is the proposed combination of deep feature detector and descriptor with ASMK matching, which outperforms existing global and local descriptors on two instance-level recognition tasks, \ie classification and search, in the domain of landmarks. 
Our ablation study shows that the proposed components reduce the memory required by ASMK.
The learned local descriptors outperform by far deep global descriptors as well as other deep local descriptors combined with ASMK. 
Finally, we provide insight into why the image-level optimization is relevant for local-descriptors and ASMK matching. 

%% file: related.tex
\section{Related work}
\label{sec:related}
We review the related work in learning global or local descriptors for instance-level matching task and local descriptors for registration tasks.

\textbf{Global descriptors.}
A common approach to obtain global image descriptors with deep fully-convolutional neural networks is to perform global pooling on 3D feature maps. This approach is applied to activations generated by pre-trained networks~\cite{bl15,tsj16,kmo16} or end-to-end learned networks~\cite{rtc19,gar+17,glj19}. One of the first examples is SPoC descriptor by Babenko and Lempitsky~\cite{bl15} that is generated by simple global sum-pooling. Weighted sum-pooling is performed in CroW by Kalantidis \etal~\cite{kmo16}, where the weights are given by the magnitude of the activation vectors at each spatial location of the feature map. Such a 2D map of weights, seen as an attention map, is related to our approach as discussed in Section~\ref{sec:relation}. Inspired by classical embeddings, Arandjelovi\'{c} \etal~\cite{agt+15} extend the VLAD~\cite{jpd+12} descriptor to NetVLAD. Its contextually re-weighted counterpart, proposed by Kim and Frahm~\cite{kdf17}, introduces a learned attention map which is generated by a small network.

\textbf{Local features and descriptors for instance-level recognition.}
Numerous classical approaches that are based on hand-crafted local features~\cite{mts+05,mcm+04} and descriptors~\cite{ms05,bet+08} exist in the literature of instance-level search~\cite{sz03,pci+07,taj16,zjs13,pls+10}. Inspired by classical feature detection, Simeoni \etal~\cite{sac19} perform MSER~\cite{mcm+04} detection on activation maps. The features detected at one feature channel are used as tentative correspondences, hence no descriptors are required; the approach is applicable to any network and does not require learning. 
Learning of attentive deep local features (DELF) is introduced in the work of Noh \etal~\cite{nas+17}. A global descriptor is derived from a network that learns to attend on feature map positions. The global descriptor is optimized with category-level labels and classification loss. At test time, locations with the strongest attention scores are selected while the descriptors are the activation vectors at the selected locations. This approach is highly relevant to ours. We therefore provide a number of different ablation experiments to reveal the key differences. 
A recent variant shows that it is possible to jointly learn DELF-like descriptors and global descriptors with a single model~\cite{cas20}. A similar achievement appears in the work of Yang \etal~\cite{ynh+20} with a scope that goes beyond instance-level recognition and covers image registration too.

\textbf{Local features and descriptors for registration.}
A richer line of work exists in learning local feature detection and description for image registration where denser point correspondences are required. As in the previous tasks, some methods do not require any learning and are applicable on any pre-trained network.
This is the case of the work Benbihi \etal~\cite{bgp19} where activation magnitudes are back-propagated to the input image and local-maxima are detected. Learning is performed with or without labeling at the local level in a number of different approaches~\cite{rwd+19,drp+19,dmr18,brp+19,bgr+20,wzh+20}. A large number of features is typically required for good performance. 
This line of research differentiates from our work; our focus is on large-scale instance-level recognition where memory requirements matter.

%% file: background.tex
\section{Background}
\label{sec:background}
In this section, the binarized versions of Selective Match Kernel (SMK) and its extension, the Aggregated Selective Match Kernel (ASMK)~\cite{taj16}\footnote{The binarized versions are originally~\cite{taj16} referred to as SMK$^\star$ and ASMK$^\star$. Only binarized versions are considered in this work and the asterisk is omitted.}, are reviewed as the necessary background. This paper exploits the ASMK indexing and retrieval.

In SMK, an image is represented by a set~$\cX = \{\vx \in \real^d\}$ of $n= |\cX|$ $d$-dimensional local descriptors. The descriptors are quantized by $k$-means quantizer $q: \real^d  \rightarrow \cC \subset \real^d$, where $\cC = \{\vc_1,\dots,\vc_k\}$ is a codebook comprising $|\cC|$ vectors (visual words). Descriptor $\vx$ is assigned to its nearest visual word $q(\vx)$. We denote by $\cX_c = \{ x \in \cX : q(x) = \vc \}$ the subset of descriptors in~$\cX$ that are assigned to visual word~$\vc$, and by 
$\cC_{\cX}$ the set of all visual words that appear in $\cX$.
Descriptor $\vx$ is mapped to a binary vector through function $b: \real^d \rightarrow \{-1,1\}^d$ given by $b(\vx) = \sign (r(\vx))$, where $r(\vx) = \vx - q(\vx)$ is the residual vector \wrt the nearest visual word and $\sign$ is the element-wise sign function.
 
The SMK similarity of two images, represented by $\cX$ and $\cY$ respectively, is estimated by cross-matching all pairs of local descriptors with match kernel 
\begin{equation}
	S_\text{\tiny SMK}(\cX,\cY) = \gamma(\cX) \, \gamma(\cY) \sum_{\vx \in \cX} \sum_{\vy \in \cY} [q(\vx)=q(\vy)] k(b(\vx) , b(\vy))	,
\label{equ:smk}
\end{equation}
where $[\cdot]$ is the Iverson bracket, $\gamma(\cX)$ is a scalar normalization that ensures unit self-similarity\footnote{To simplify, we use the same notation, \ie $\gamma(\cdot)$, for the normalization of different similarity measures in the rest of the text. In each case, it ensures unit self-similarity of the corresponding similarity measure.}, \ie $S_\text{\tiny SMK}(\cX,\cX)=1$.
Function $k: \{-1,1\}^d \times \{-1,1\}^d \rightarrow [0,1]$ is given by 
\begin{equation}
k(b(\vx), b(\vy)) = 
\left\{
		\begin{array}{r@{\lsp}l}
			\left(\frac{b(\vx)^\top b(\vy)}{d}\right)^\alpha, & \frac{b(\vx)^\top b(\vy)}{d} \geq \tau \\
			0,                     & \text{otherwise},
		\end{array}
\right.
\label{equ:local-kernel}
\end{equation}
where $\tau \in [0,1]$ is a threshold parameter.
Only descriptor pairs that are assigned to the same visual word contribute to the image similarity in \equ{smk}. In practice, not all pairs need to be enumerated and image similarity is equivalently given by 
\begin{equation}
	S_\text{\tiny SMK}(\cX,\cY) = \gamma(\cX) \, \gamma(\cY) \sum_{\vc \in \cC_{\cX} \cap \cC_{\cY}} \sum_{\vx \in \cX_{\vc}} \sum_{\vy \in \cY_{\vc}} k(b(\vx) , b(\vy))	,
\label{equ:smk-fast}
\end{equation}
where cross-matching is only performed within common visual words. 

ASMK first \emph{aggregates} the local descriptors assigned to the same visual word into a single binary vector. This is performed by $B(\cX_c) = \sign \left(\sum_{x\in \cX_c} r(\vx)\right)$, with $B(\cX_c)\in \{-1,1\}^d$. Image similarity in ASMK is given by 
\begin{equation}
	S_\text{\tiny ASMK}(\cX,\cY) = \gamma(\cX) \, \gamma(\cY) \sum_{\vc \in \cC_{\cX} \cap \cC_{\cY}} k(B(\cX_c) , B(\cY_c)).
\label{equ:asmk}
\end{equation}
This is computationally and memory-wise more efficient than SMK. In practice, it is known to perform better due to handling the burstiness phenomenon~\cite{jds09}.
Efficient search is performed by using an inverted-file indexing structure.

\textbf{Simplifications.} Compared to the original approach~\cite{taj16}, we drop IDF weighting, pre-binarization random projections, and median-value thresholding, as these are found unnecessary.

%% file: method.tex
\section{Method}
\label{sec:method}
Learning local descriptors with ASMK in an end-to-end manner is challenging and impractical due to the use of large visual codebooks and the hard-assignment of descriptors to visual words.
In this section, we first describe a simple framework to generate global descriptors that can be optimized with image-level labels and provide an insight into why this is relevant to optimizing the local representation too. Then, we extend the framework by additional components and discuss their relation to prior work.

\subsection{Derivation of the architecture} \label{sec:architecture}
In the following, let us assume a deep Fully Convolutional Network (FCN), denoted by function
$f: \real^{w \times h \times 3} \rightarrow  \real^{W \times H \times D}$, that maps an input image $I$ to a 3D tensor of activations $f(I)$. 
The FCN is used as an extractor of dense deep local descriptors.
The 3D activation tensor can be equivalently seen as a set $\cU  = \{\vu \in \real^D\} $ of $W \times H$ $D$-dimensional local descriptors\footnote{Both $f(I)$ and $\cU$ correspond to the same representation seen as a 3D tensor and a set of descriptors, respectively. We write $\cU=f(I)$ implying the tensor is transformed into a set of vectors. $\cU$ is, in fact, a multi-set, but it is referred to as set in the paper.}.
Each local descriptor is associated to a keypoint, also called local feature, that is equivalent to the receptive field, or a fraction of it, of the corresponding activations. 

Let us consider global image descriptors constructed by global sum-pooling, known as SPoC~\cite{bl15}. Pairwise image similarity is estimated by the inner product of the corresponding \l2-normalized SPoC descriptors. This can be equivalently seen as an efficient match kernel.
Let $\cU = f(I)$ and $\cV = f(J)$ be sets of dense feature descriptors in images $I$ and $J$, respectively. Image similarity is given by
\begin{align}
	S_\text{\tiny SPoC}(\cU,\cV) &= \gamma(\cU) \, \gamma(\cV) \sum_{\vu \in \cU} \sum_{\vv \in \cV}  \vu^\top \vv \label{equ:simple-kernel-1}\\
	&= \left(\gamma(\cU) \sum_{\vu \in \cU} \vu \right)^\top  \left(\gamma(\cV) \sum_{\vv \in \cV} \vv \right) = Z_\text{\tiny SPoC}(\cU)^\top Z_\text{\tiny SPoC}(\cV)\mbox{,} \label{equ:simple-kernel-2}
\end{align}
where $\gamma(\cU) = 1 / ||\sum_{\vu \in \cU} \vu||$. The optimization of the network parameters is cast as metric learning.

The interpretation of global descriptor matching in \equ{simple-kernel-2} as local descriptor cross-matching in \equ{simple-kernel-1} provides some useful insight. Local descriptor similarity is estimated by $\vu^\top\vv = ||\vu|| \cdot ||\vv|| \cdot \cos(\vu,\vv)$.
Optimizing SPoC with image-level labels and contrastive loss implicitly optimizes the following four cases.
First, local descriptors of background features, or image regions, are pushed to have small magnitude, \ie small $||\vu||$, so that their contribution in cross-matching is minimal.
Similarly, local descriptors of foreground features are pushed to have large magnitude.
Additionally, local descriptors of truly-corresponding features, \ie image locations depicting the same object or object part, are pushed closer in the representation space, so that their inner product is maximized.
Finally, local descriptors of non-corresponding features are pushed apart, so that their inner product is minimized.
Therefore, optimizing global descriptors is a good surrogate to optimize local descriptors that are used to measure similarity via efficient match kernels, such as in ASMK.
Importantly, this is possible with image-level labels and no local correspondences are required for training. 
In the following, we introduce additional components to the presented model, that are designed to amplify these properties.

\begin{figure}[t]
\centering
\includegraphics[width=0.99\linewidth]{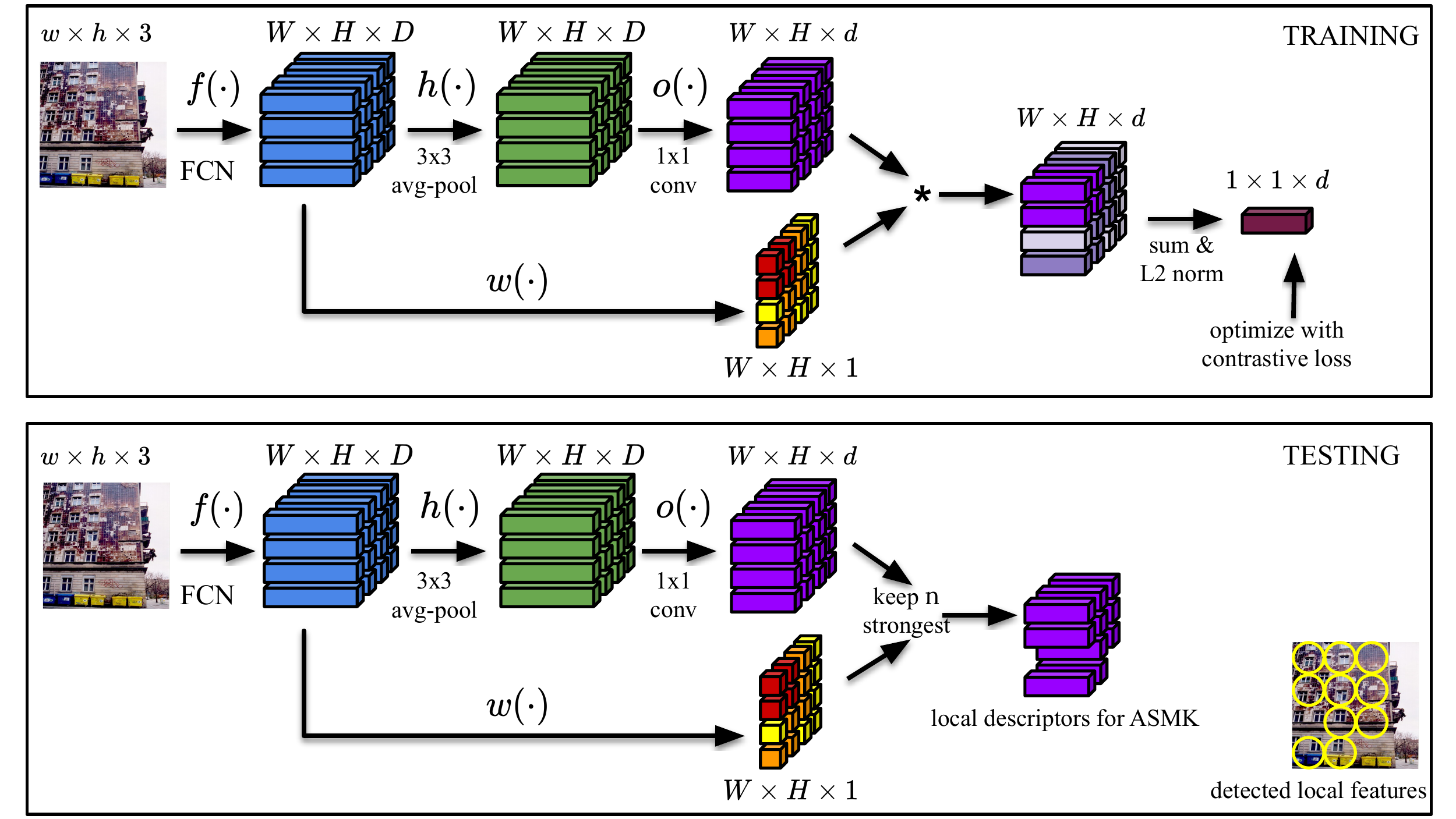}
\vspace{-10pt}
\caption{Training and testing architecture overview for HOW local features and descriptors. A global descriptor is generated for each image during training and optimized with contrastive loss and image-level labels. During testing the strongest local descriptors (features), according to the attention map, are kept to represent the image. Then, these are used with ASMK for image search.\label{fig:teaser}
\vspace{-10pt}
}
\end{figure}

\textbf{Feature strength and attention.}
The feature strength, or importance, is estimated as the \l2 norm of the feature descriptor $\vu$ by the attention function  $w(\vu) = ||\vu||$.
During training, the feature strength is used to weigh the contribution of each feature descriptor in cross-matching.
This way, the impact of the weak features is limited during the training, which is motivated by only the strongest features being selected in the test time.
The attention function is fixed, without any parameters to be learned.

\textbf{Local smoothing.} Large activation values tend to appear in multiple channels of the activation tensor on foreground features~\cite{sac19}. Moreover, these activations tend not to be spatially well aligned. 
We propose to spatially smooth the activations by average pooling in an $M\times M$ neighborhood. The result is denoted by $\bar{\cU}=h(f(I))$ or $\bar{\cU}=h(\cU)$. 
Our experiments show that it is beneficial for the aggregation operation performed in ASMK. It is a fixed function, without any further parameters to be learned, and parameter $M$ is a design choice.

\textbf{Mean subtraction.} Commonly used FCNs (all that are used in this work) generate non-negative activation tensors since a Rectified Linear Unit (ReLU) constitutes the last layer. Therefore, the inner product for all local descriptor pairs in cross-matching is non-negative and contributes to the image similarity. Mean descriptor subtraction is known to capture negative evidence~\cite{jc12} and allows to better disambiguate non-matching descriptors.

\textbf{Descriptor whitening.} Local descriptor dimensions are de-correlated by a linear whitening transformation; PCA-whitening improves the discriminability of both local~\cite{blv+17} and global descriptors~\cite{rsa+14}. For efficiency, dimensionality reduction is performed jointly with the whitening. Formally, we group mean subtraction, whitening, and dimensionality reduction into function $o: \real^D \rightarrow \real^d$ given by $o(\vu) = P(\vu-\vm)$, where $P \in \real^{d\times D}$, $d \leq D$. Function $o(\cdot)$ is implemented by $1\times 1$ convolution with bias. In practice, we initialize $P$ and $\vm$ according to the result of PCA whitening on a set of local descriptors from the training set and keep them fixed during the training.

\textbf{Learning.} Let $\bar{\cV} = h(\cV)$ and $\bar{\vv} \in \bar{\cV}$ denote the activation vector after local average pooling $h(\cdot)$ at the same spatial location as $\vv \in \cV$. Similarly for $\bar{\cU}$ and $\bar{\vu}$. 
The image similarity that is being optimized during learning is expressed as
\begin{align}
	S_\ghw(\cU,\cV) &= \gamma(\bar{\cU}) \, \gamma(\bar{\cV}) \sum_{\vu \in \cU} \sum_{\vv \in \cV} w(\vu) \cdot  w(\vv) \cdot o(\bar{\vu})^\top o(\bar{\vv}) \label{equ:extended-kernel-1}\\
	&= \left(\!\gamma(\bar{\cU})\! \sum_{\vu \in \cU} w(\vu) o(\bar{\vu})\!\!\right)^{\!\!\!\top}\!\!  
	  \left(\!\gamma(\bar{\cV})\! \sum_{\vv \in \cV} w(\vv) o(\bar{\vv})\!\!\right) = Z_{\ghw}(\cU)^{\!\top} Z_{\ghw}(\cV)\mbox{.}\label{equ:extended-kernel-2}
\end{align}
A metric learning approach is used to train the network. In particular, contrastive loss is minimized: $||Z_{\ghw}(\cU)-Z_{\ghw}(\cV))||^2$ if $\cU$ and $\cV$ originate from matching (positive) image pairs, or $([\mu-||Z_{\ghw}(\cU)-Z_{\ghw}(\cV)||]_{+})^2$ otherwise, where $[\cdot]_{+}$ denotes the positive part.

\begin{figure}
\vspace{-10pt}
\centering
\begin{tabular}{cccc}
\raisebox{30pt}{\multirow{2}{*}{\includegraphics[width=.33\linewidth]{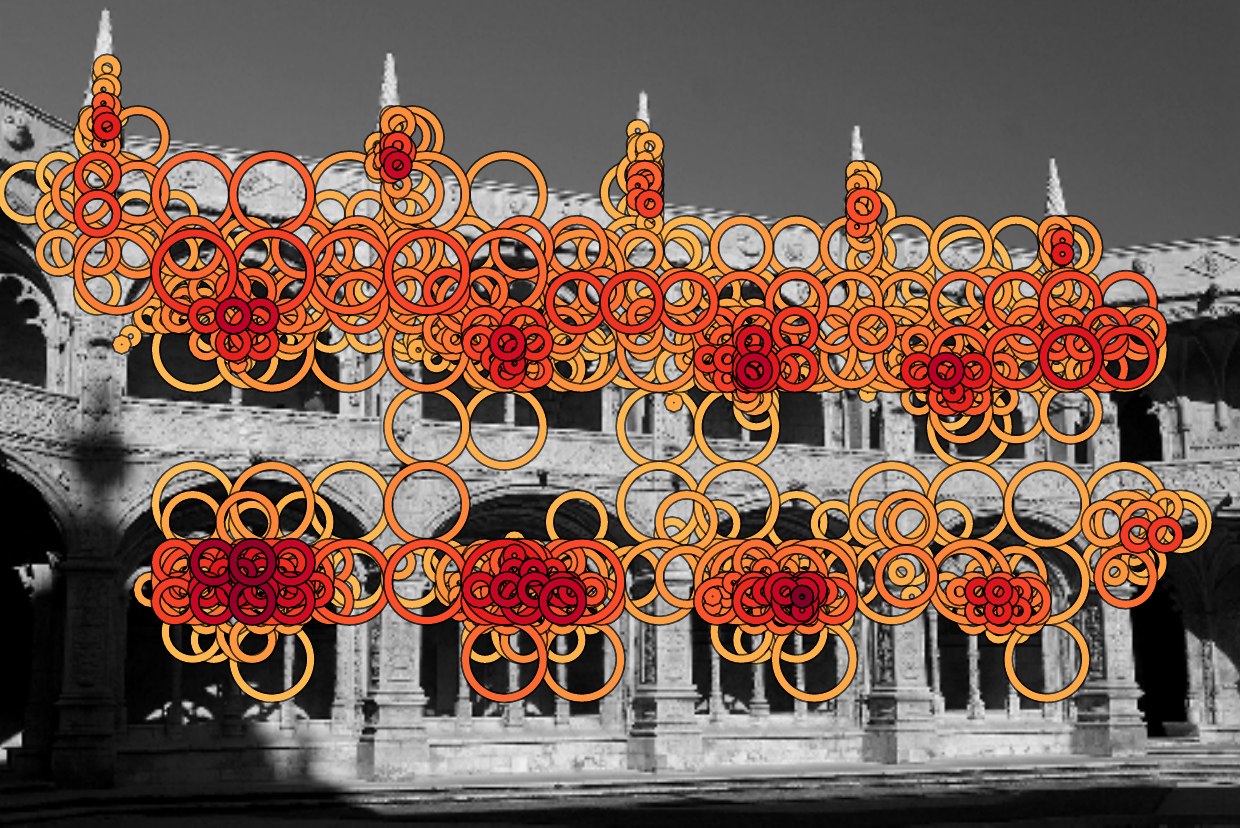}}}&
\raisebox{30pt}{\multirow{2}{*}{\includegraphics[width=.33\linewidth]{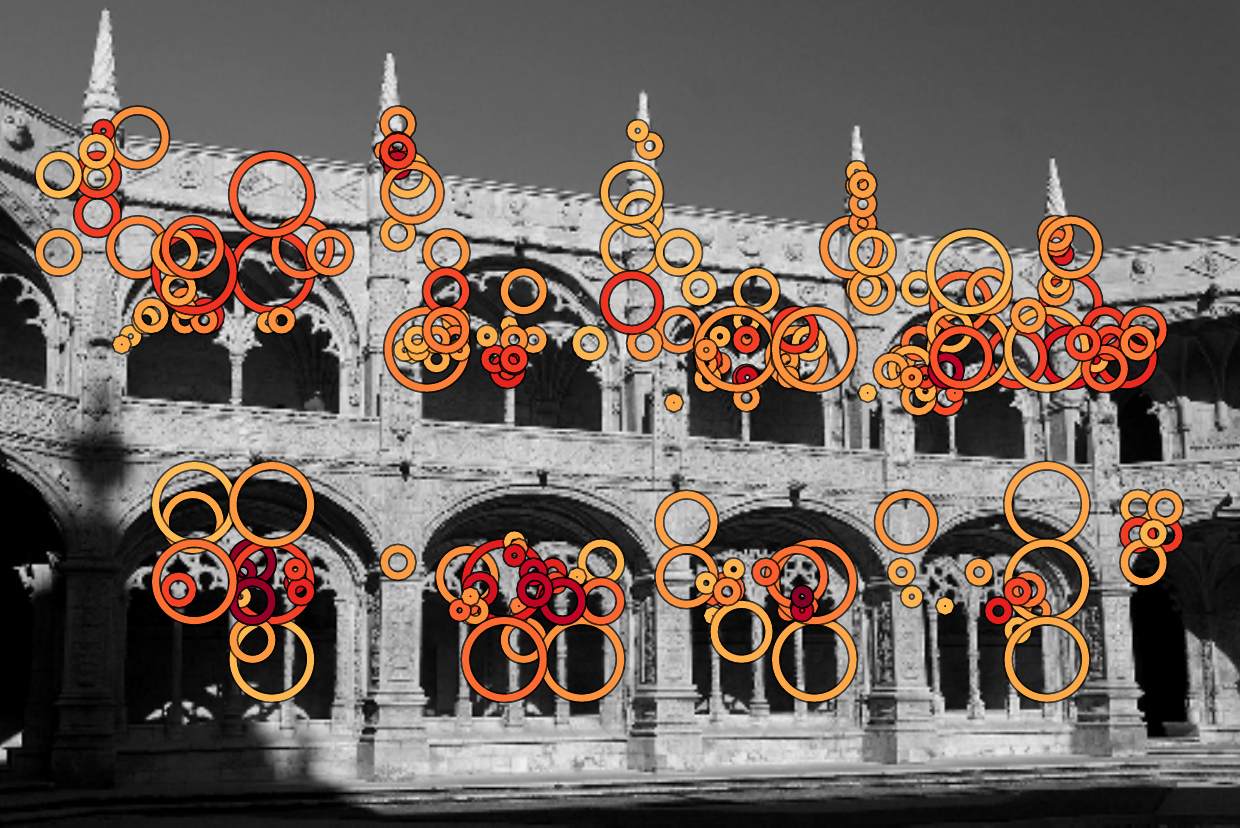}}}& 
\includegraphics[width=.15\linewidth]{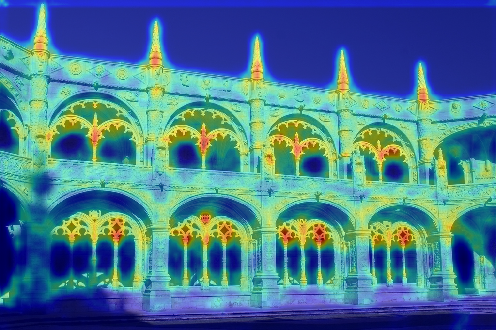}
& \includegraphics[width=.15\linewidth]{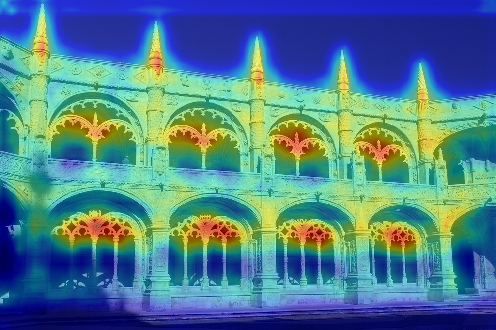}\\
& & \includegraphics[width=.15\linewidth]{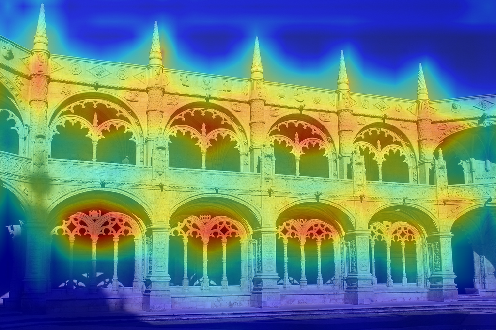}
& \includegraphics[width=.15\linewidth]{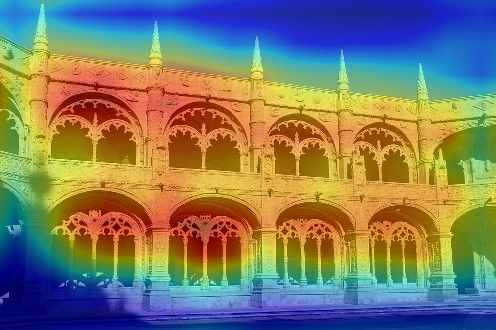} \\
%
%
\end{tabular}
\vspace{-10pt}
\caption{Example of multi-scale local feature detection. Left: Strongest 1,000 local features with color-coded strength; red is the strongest. Middle: Only the strongest feature per visual word is shown. Right:  Attention maps for input images resized by scaling factors 0.25, 0.5, 1, and 2.\label{fig:example}
\vspace{-15pt}}
\end{figure}

\textbf{Test-time architecture.}
The architecture in test time is slightly modified; no global descriptor is generated.
The $n$ strongest descriptors $o(\bar{\vu})$ for $\vu \in \cU$ are kept according to the importance given by $w(\vu)$.
These are used as the local descriptor set $\cX$ in ASMK.
Multi-scale extraction is performed by resizing the input image at multiple resolutions. Local features from all resolutions are merged and ranked jointly according to strength. Selection of the strongest features is performed in the merged set. 
Multi-scale extraction is performed only during testing, and not during training.
The training and testing architectures are summarized in Figure~\ref{fig:teaser}, while examples of detected features are shown in Figure~\ref{fig:example}.

\subsection{Relation to prior work}
\label{sec:relation}
The proposed method has connections to different approaches in the literature which are discussed in this section.
The work closest to ours, in terms of training of the local detector and descriptor, is DELF~\cite{nas+17}. Following our notation, DELF generates local feature descriptors $Z_{\scriptscriptstyle \hatghw}(\cU)$, where $\hat{h}(.)$ is identity, \ie no local smoothing, $\hat{o}(.)$ is identity, \ie no mean subtraction, whitening, and dimensionality reduction, and $\hat{w}(.)$ is a learned attention function. In particular, $\hat{w}: \real^D \rightarrow \real_{+}$ is a 2 layer convolutional network with $1\times 1$ convolutions whose parameters are learned during the training. Dimensionality reduction of the descriptor space in DELF is performed as post-processing and is not a part of the optimization.\footnote{The main difference is that we do not follow the two stage training performed in the original work~\cite{nas+17}; DELF is trained in a single stage for our ablations.}
In terms of optimization, DELF performs the training in a classification manner with cross entropy loss. 
We show experimentally, that when combined with ASMK, the proposed descriptors are superior to DELF.

\textbf{Fixed attention.}
Function $w(\cdot)$ is previously used to weigh activations and generate global descriptor, in particular by CroW~\cite{kmo16}. It is also used in concurrent work for deep local features by Yang \etal~\cite{ynh+20}. The same attention function is used by Iscen \etal~\cite{itg+15} for feature detection on top of dense SIFT descriptors without any learning. In our case, during learning, background or domain irrelevant descriptors are pushed to have low \l2 norm in order to contribute less. The corresponding features are consequently not detected during test time.

\textbf{Learned attention.}
Further example of learned attention, apart from DELF, is the contextual re-weighting that is performed to construct global descriptors in the work of Kim and Frahm~\cite{kdf17}. 
The attention function is similar to DELF but with larger spatial kernel; a contextual neighborhood is used.
 A comparison between learned and fixed attention is included in the experimental ablations in Section~\ref{sec:exp}.

\textbf{Whitening.}
A common approach to whitening is to apply it as the last step in the pipeline, as post-processing.
A similar approach to ours -- applying the PCA whitening during training and learning it end-to-end -- is followed by Gordo \etal~\cite{gar+17} in the context of processing and aggregating regional descriptors to construct global descriptors.
Comparison between ``in-network'' and post-processing whitening-reduction is included in Section~\ref{sec:exp}.

%% file: experiments.tex
\section{Experiments}
\label{sec:exp}
We first review the datasets used for training, validation, and testing. Then, we discuss the implementation details for training and for testing with ASMK. Finally, we present our results on two instance-level tasks, namely recognition and search in the domain of landmarks and buildings.

\subsection{Datasets}

\textbf{Training. } The training dataset \emph{SfM120k}~\cite{rtc19} is used. It is the outcome of Stucture-from-Motion (SfM)~\cite{src+15} with 551 3D models for training. Matching pairs (anchor-positive) are formed by images with visual overlap (same 3D model). Non-matching pairs (anchor-negative) come from different 3D models.

\textbf{Validation. } We use the remaining 162 3D models of SfM120k to construct a challenging validation set reflecting the target task; this is different than the validation in~\cite{rtc19}.
We randomly choose 5 images per 3D model as queries. Then, for each query, images of the same 3D model with enough (more than 3), but not too many (at most 10), common 3D points with the query are marked as positive images to be retrieved. This avoids dominating the evaluation measure by a large number of easy examples. The remaining images of the same 3D model are excluded from evaluation for the specific query~\cite{pci+07}. Skipping queries with an empty list of positive images results in 719 queries and 12,441 database images in total. Evaluation on the validation set is performed by instance-level search and performance is measured by mean Average Precision (mAP). 

\textbf{Evaluation on instance-level search.}
We use \roxf~\cite{pci+07} and \rpar~\cite{pci+08} to evaluate search performance in the revisited benchmark~\cite{rit+18}. They consist of 70 queries each, and 4993 and 6322 database images, respectively, and 1 million distractors called \rdis.
We measure mAP on the Medium and Hard setups.

\textbf{Evaluation on instance-level classification. }
We use instance-level classification as another task on which to evaluate the performance of the learned local descriptors.
We perform search with ASMK and use k-nearest-neighbors classifiers for class predictions.
The \emph{Google Landmarks Dataset -- version 2} (GLD$_2$)~\cite{wac+20} is used. It consists of 4,132,914 train/database images with known class labels, and 117,577 test/query images which either correspond to the database landmarks, to other landmarks, or to non-landmark images. The query images are split into testing (private) and validation (public) sets with 76,627 and 40,950 images, respectively. Performance is measured by micro Average Precision ($\mu$AP)~\cite{plr09}, also known as Global Average Precision (GAP), on the testing split. 
Note that we do not perform any learning on this dataset.
We additionally create a mini version of GLD$_2$ to use for ablation experiments. It includes 1,000 query images, that are sampled from the testing split, and 10,0000 database images with labels, where the images come from 50 landmarks in total. We denote it by Tiny-GLD$_2$.

Classification is performed by accumulating the $N$ top-ranked images per class. Prediction is given by the top-ranked class and the corresponding confidence is equal to the accumulated similarity. We use three variants, \ie $N=1$ (CLS1), $N=10$ (CLS2), and $N=10$ with accumulation of square-rooted similarity multiplied by a class weight (CLS3). The class weight is equal to the logarithm of the number of classes divided by the class frequency to down-weigh frequent classes. 

\subsection{Implementation details}

\textbf{Network architecture.} We perform experiments with a backbone FCN ResNet18 and ResNet50, initialized by pre-training on ImageNet~\cite{rds+15}. 
Descriptor dimensionality $D$ is equal to 512 and 2048, respectively. We additionally experiment by removing the last block, \ie ``conv5\_x'', where $D$ becomes 256 and 1024, respectively.
We set $d=128$ and $M=3$ to perform $3\times 3$ average pooling for local smoothing.

\textbf{Training. }
We use a batch size of 5 tuples, where a tuple consists of 7 images -- an anchor, a positive, and 5 negatives.
Training images are restricted to a maximum resolution of 1,024 pixels.
For each epoch, we randomly choose 2,000 anchor-positive pairs.
The pool of candidate negatives contains 20,000 randomly chosen images, and hard-negative mining is performed before every epoch. 
We adopt the above choices from the work of Radenovic \etal~\cite{rtc19}, whose public implementation\footnote{\url{https://github.com/filipradenovic/cnnimageretrieval-pytorch}} we use to implement our method.	
To initialize $P$ and $\vm$, we use local descriptors from 5,000 training images extracted at a single scale.
We use Adam optimizer with weight decay equal to $10^{-4}$.
The learning rate and margin $\mu$ are tuned, per variant, according to the performance on the validation set, by trying values $10^{-6}, 5\cdot 10^{-6}, 10^{-5},5\cdot 10^{-5}$ for learning rate and $0.5$ to $0.9$ with step $0.05$ for margin $\mu$. 
Margin $\mu$ is set equal to $0.8$ for the proposed method, and learning rate equal to $5\cdot 10^{-6}$ and $10^{-5}$ for ResNet18 and ResNet50 without the last block, respectively, according to the tuning process.
Training is performed for 20 epochs and 1 epoch takes about 22 minutes for ResNet50 without the last block on a single NVIDIA Tesla V100 GPU with 32GB of DRAM.
Training with cross entropy loss for ablations is performed with a batch size equal to 64 for 10 epochs.
We repeat the training of each model and report mean and standard deviation over 5 runs. In large scale experiments, we evaluate a single model, the one with median performance on the validation set.

\textbf{Validation. }
Validation performance is measured with ASMK-based search on the validation set. We measure validation performance every 5 epochs during training and the best performing model is kept. 

\textbf{Testing. }
We use ASMK to perform testing and to evaluate the performance of the learned local descriptors. 
The default ASMK configuation is as follows. 
We set threshold $\tau=0$, $d=128$, and use a codebook of $\kappa=65536$ visual words, which is learned on local descriptors from 20,000 training images extracted at a single scale. 
Images are resized to have maximum resolution of 1024 pixels and multi-scale extraction is performed by re-scaling with factors 0.25, 0.353, 0.5, 0.707, 1.0, 1.414, 2.0. Assignment to multiple, in particular 5, visual words is performed for the query images. The strongest $n=1000$ local descriptors are kept.
The default configuration is used unless otherwise stated. The inverted file is compressed by delta coding.

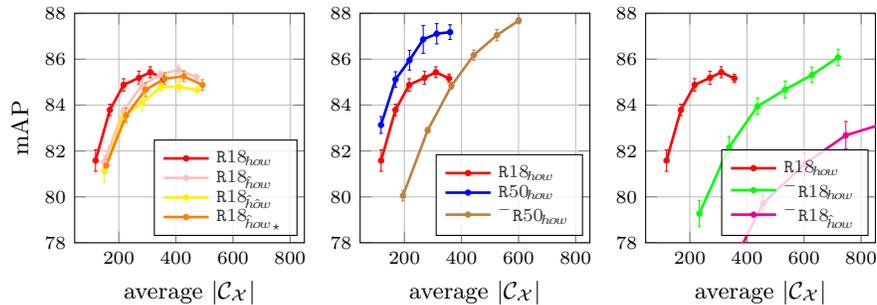
\begin{figure}[t]
\input{fig/perf_vs_mem}
\caption{Ablation study reporting performance versus average number of vectors per image in ASMK on the validation set for varying $n$ (400,600,800,1000,1200,1400). Components used by DELF are denoted with ~$\hat{}$~, while ours without. $\star$: random initialization of $\vm$ and $P$  (learned during training). Mean and standard deviation over 5 runs.
\label{fig:abla}
}
\end{figure}

\subsection{Ablation experiments}
We denote ResNet18 and ResNet50 by \resnet{18}{}{} and \resnet{50}{}{}, and their versions with the last block skipped by \resnet{18}{-}{} and \resnet{50}{-}{}, respectively.
The local smoothing, whitening with reduction, and the fixed attention are denoted by subscripts ~$h$, $o$ and $w$, respectively. The proposed method is denoted by \resnet{18}{}{\ghw} when the backbone network is ResNet18. Following the same convention, the original DELF architecture is denoted by \resnet{50}{-}{\hatghw}, where the dimensionality reduction is not part of the network but is performed by PCA whitening as post-processing.	

Figure~\ref{fig:abla} shows the performance on the validation set versus the average number of binary vectors indexed by ASMK for the database images. 
We perform an ablation by excluding the proposed components and by using different backbone networks. 
Local smoothing results in larger amount of aggregation in ASMK (red vs pink) and reduces the memory requirements, which are linear in $|\cC_{\cX}|$. It additionally improves the performance when the last ResNet block is removed and feature maps have two times larger resolution (green vs magenta). 
Initializing and fixing $h(\cdot)$ with the result of PCA whitening is a beneficial choice too (orange vs pink). ResNet50 gives a good performance boost compared to ResNet18 (blue vs red), while removing the last block is able to reach higher performance at the cost of larger memory requirements (brown vs blue). 
More ablation experiments are shown in Table~\ref{tab:abla}. 
Fixed attention is better than learned attention (3 vs 4). 
Metric learning with the contrastive loss delivers significantly better performance than cross entropy loss, in a classification manner, as done by Noh \etal~\cite{nas+17} (4 vs 5). This is a confirmation of results that appear in the literature of instance-level recognition~\cite{gar+17,vjh17}.

\begin{table}[t]
\input{tab/ablations}
\caption{Ablation study for performance and average number of descriptors per image in ASMK (mean and standard deviation over 5 runs).
1: our method, 5: DELF variant.
CO: contrastive loss, CE: cross entropy. On Tiny-GLD$_2$, classifier CLS3 is used.
\label{tab:abla}
}
\end{table}
\begin{figure}[b!]
\input{fig/train_curve}
\caption{Training loss and validation performance during the training. Mean and standard deviation ($\times$5 for better visualization) is reported over 5 runs. Both performance curves correspond to the same model, only its inference differs.
\label{fig:train}
}
\end{figure}
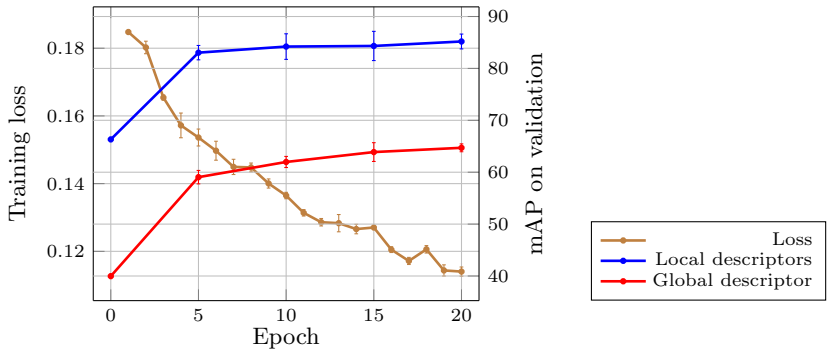

In Figure~\ref{fig:train}, we present the evolution of the model during training. 
We evaluate the performance of the optimized global descriptor for nearest neighbor search with multi-scale global descriptors, \ie aggregation of global descriptors extracted at multiple image resolutions (same set of 7 resolutions as for the local descriptors).
We additionally evaluate performance of the corresponding local descriptors with ASMK.
The descriptor that is directly optimized in the loss performs worse than the internal local descriptors.

In the following we use two backbone networks -- \resnet{18}{}{}, which is fast with low memory footprint, and \resnet{50}{-}{}, which achieves better results at the cost of slower extraction and more memory.

\begin{table}[t]
\input{tab/rop}
\caption{Performance comparison with global and local descriptors for instance-level search on \roxf (\ro) and \rpar (\rp).
Memory is reported for \rdis. Methods marked by $\dagger$ are evaluated by us using the public models for descriptor extraction.
The method marked by $\ddagger$ is evaluated by us using the public descriptors~\cite{rit+18}. GEM$_w$ is a public model that includes a ``whitening'' (FC) layer. Dimensionality reduction is denoted by $\triangleright$ and descriptor concatenation by $+$.
PQ8 and PQ1 denote PQ quantization using 8D and 1D sub-spaces, respectively.
\label{tab:soasearch}}
\end{table}

\subsection{Large-scale instance-level search}
The performance comparison on \roxf and \rpar is presented in Table~\ref{tab:soasearch}. We do our best to evaluate the best available variants or models of the state-of-the-art approaches. The proposed local descriptors and DELF descriptors are evaluated with identical implementation and configuration of ASMK. The proposed descriptors outperform all approaches by a large margin at large scale; global descriptors perform well enough at small scale, but at large scale, local representation is essential. 
Even with a backbone network as small as ResNet18, we achieve the second best performance (ranked after our method with ResNet50) on all cases of \roxf and at the hard setup of \rpar+\rdis.
Compared to DELF, our descriptors, named \emph{HOW}, perform better for less memory.

Teichmann \etal~\cite{taz+19} achieve average performance (mean all in Table~\ref{tab:soasearch}) equal to 56.0 and 57.3 without and with spatial verification, respectively. They use additional supervision, \ie manually created bounding boxes for 94,000 images, which we do not.
The concurrent work of Cao \etal~\cite{cas20} achieves average performance equal to 58.3 but requires 485 GB of RAM, and slightly lower performance with 22.6 GB of RAM for binarized descriptors.

In an effort to further compress the memory requirements of competing global descriptors, we evaluate the best variant with dimensionality reduction and with Product Quantization (PQ)~\cite{jds11}.
We further improve their performance by concatenating the two best performing ones. 
Among all these variants, the proposed approach appears to be a good solution in the performance-memory trade-off. 

Storing less vectors in ASMK affects memory but also speed. A query of average statistics computes the hamming distance for about $1.2\cdot 10^6$ (average number of vectors stored per inverted list multiplied by average $|\cC_{\cX}|$) 128D binary vector pairs in the case of \resnet{18}{-}{} with our method at large scale. The same number for \resnet{50}{-}{} is $3.2\cdot10^6$. Search on \roxf+\rdis takes on average 0.75 seconds on a single threaded Python non-optimized CPU implementation.

\begin{table}[t!]
\input{tab/gl}
\caption{Performance comparison on instance-level recognition (GLD$_2$). 
$\mu$AP is reported for classification with 3 different k-nn classifiers. 
Existing methods are evaluated by us using the public models.
Global descriptors are combined with simple nearest neighbor search, and local descriptors are combined with ASMK-based retrieval.
\label{tab:soaclass}
}
\end{table}

\subsection{Large-scale instance-level classification}
Performance comparison on GLD$_2$ is presented in Table~\ref{tab:soaclass}.
We extract DELF keeping at most top 1,000 local descriptors and reduce dimensionality to 128. The proposed local descriptors and DELF descriptors are evaluated with identical configuration of ASMK.
Our method outperforms all other methods with memory requirements that are even lower than raw 2048D global descriptors. 
DELF, \resnet{18}{}{\ghw}, and \resnet{50}{-}{\ghw}, end up with 347, 252, and 423 vectors to store per image on average, respectively.
Due to the strength threshold, DELF extracted 504 local descriptors per image on average which is significantly less than for images of \rdis.
Multiple visual word assignment is not performed in this experiment, for any of the methods, to reduce the computational cost of search.

%% file: fig/perf_vs_mem.tex
\input{fig/data}
\begin{tikzpicture}
\begin{axis}[%
	width=0.38\linewidth,
	height=0.38\linewidth,
	xlabel={\small average $|\mathcal{C}_{\mathcal{X}}|$},
	ylabel={\small mAP},
	legend cell align={left},
	legend pos=south east,
	legend style={cells={anchor=west}, font =\scriptsize, fill opacity=0.8, row sep=-2.5pt},
	xmax = 100,
	xmax = 850,
	ymin = 78,
	ymax = 88,
	grid=both,
  error bars/y explicit = true,
]
	\addplot[color=red,     solid, mark=*,  mark size=0.7, line width=1.0, error bars/.cd, error mark=-,y dir=both] table[x=mem, y expr={\thisrow{sco}}, y error expr={\thisrow{scostd}}] \redcurve;\leg{\resnet{18}{}{\!h\!o\!w}};
	\addplot[color=pink,     solid, mark=*,  mark size=0.7, line width=1.0, error bars/.cd, error mark=-,y dir=both] table[x=mem, y expr={\thisrow{sco}}, y error expr={\thisrow{scostd}}] \pinkcurve;\leg{\resnet{18}{}{\!\hat{h}\!o\!w}};
	\addplot[color=yellow,     solid, mark=*,  mark size=0.7, line width=1.0, error bars/.cd, error mark=-,y dir=both] table[x=mem, y expr={\thisrow{sco}}, y error expr={\thisrow{scostd}}] \yellowcurve;\leg{\resnet{18}{}{\!\hat{h}\!\hat{o}\!w}};
	\addplot[color=orange,     solid, mark=*,  mark size=0.7, line width=1.0, error bars/.cd, error mark=-,y dir=both] table[x=mem, y expr={\thisrow{sco}}, y error expr={\thisrow{scostd}}] \orangecurve;\leg{\resnet{18}{}{\!\hat{h}\!o\!w_\star}};
	\end{axis}
\end{tikzpicture}
\begin{tikzpicture}
\begin{axis}[%
	width=0.38\linewidth,
	height=0.38\linewidth,
	xlabel={\small average $|\mathcal{C}_{\mathcal{X}}|$},
	legend cell align={left},
	legend pos=south east,
	legend style={cells={anchor=west}, font =\scriptsize, fill opacity=0.8, row sep=-2.5pt},
	xmax = 100,
	xmax = 850,
	ymin = 78,
	ymax = 88,
	grid=both,	
	error bars/y explicit = true,
]
	\addplot[color=red,     solid, mark=*,  mark size=0.7, line width=1.0, error bars/.cd, error mark=-,y dir=both] table[x=mem, y expr={\thisrow{sco}}, y error expr={\thisrow{scostd}}] \redcurve;\leg{\resnet{18}{}{\!h\!o\!w}};
\addplot[color=blue,     solid, mark=*,  mark size=0.7, line width=1.0, error bars/.cd, error mark=-,y dir=both] table[x=mem, y expr={\thisrow{sco}}, y error expr={\thisrow{scostd}}] \bluecurve;\leg{\resnet{50}{}{\!h\!o\!w}};
\addplot[color=brown,     solid, mark=*,  mark size=0.7, line width=1.0, error bars/.cd, error mark=-,y dir=both] table[x=mem, y expr={\thisrow{sco}}, y error expr={\thisrow{scostd}}] \browncurve;\leg{\resnet{50}{-}{\!h\!o\!w}};
\end{axis}
\end{tikzpicture}
\begin{tikzpicture}
\begin{axis}[%
	width=0.38\linewidth,
	height=0.38\linewidth,
	xlabel={\small average $|\mathcal{C}_{\mathcal{X}}|$},
	legend cell align={left},
	legend pos=south east,
	legend style={cells={anchor=west}, font =\scriptsize, fill opacity=0.8, row sep=-2.5pt},
	xmax = 100,
	xmax = 850,
	ymin = 78,
	ymax = 88,
	grid=both,
  error bars/y explicit = true,
]
	\addplot[color=red,     solid, mark=*,  mark size=0.7, line width=1.0, error bars/.cd, error mark=-,y dir=both] table[x=mem, y expr={\thisrow{sco}}, y error expr={\thisrow{scostd}}] \redcurve;\leg{\resnet{18}{}{\!h\!o\!w}};
	\addplot[color=green,     solid, mark=*,  mark size=0.7, line width=1.0, error bars/.cd, error mark=-,y dir=both] table[x=mem, y expr={\thisrow{sco}}, y error expr={\thisrow{scostd}}] \greencurve;\leg{\resnet{18}{-}{\!h\!o\!w}};
  \addplot[color=magenta,     solid, mark=*,  mark size=0.7, line width=1.0, error bars/.cd, error mark=-,y dir=both] table[x=mem, y expr={\thisrow{sco}}, y error expr={\thisrow{scostd}}] \magentacurve;\leg{\resnet{18}{-}{\!\hat{h}\!o\!w}};
\end{axis}
\end{tikzpicture}

%% file: tab/ablations.tex
%
\begin{center}
\def\arraystretch{0.8}
\begin{tabular}	{l@{\msp}c@{\lsp}c@{\xssp}r@{\lsp}c@{\xssp}r@{\lsp}c@{\xssp}r@{\lsp}c@{\xssp}r@{\lsp}c@{\xssp}r@{\lsp}c@{\xssp}r}
\hline
\multirow{2}{*}{Method}                              &   \multirow{2}{*}{Loss}             & \multicolumn{4}{c}{Validation} & \multicolumn{4}{c}{\roxf} & \multicolumn{4}{c}{Tiny-GLD$_2$}       \\ \cline{3-14}
  											                             &       					&       \multicolumn{2}{c}{mAP}    &    \multicolumn{2}{c}{$|\cC_{\cX}|$}    &      \multicolumn{2}{c}{mAP}   &  \multicolumn{2}{c}{$|\cC_{\cX}|$} &     \multicolumn{2}{c}{$\mu$AP}      &      \multicolumn{2}{c}{$|\cC_{\cX}|$}  \\ \hline \hline
\blue{1:~\resnet{18}{}{\ghw}}     & \blue{CO}   & \blue{85.2{\tiny$\pm$}} &\blue{{\tiny 0.3}} & \blue{263.8{\tiny$\pm$}} &\blue{{\tiny 0.1}} &  \blue{74.8{\tiny$\pm$}} &\blue{{\tiny 0.2}} & \blue{283.6{\tiny$\pm$}} &\blue{{\tiny 0.2}} &  \blue{81.3{\tiny$\pm$}} &\blue{{\tiny 1.0}} & \blue{252.2{\tiny$\pm$}} &\blue{{\tiny 0.5}}  \\               
2:~\resnet{18}{}{\ioo}	 			    & CO          & 85.3{\tiny$\pm$} &{\tiny 0.2} & 344.1{\tiny$\pm$} &{\tiny 0.7} &  75.4{\tiny$\pm$} &{\tiny 0.3} & 365.9{\tiny$\pm$} &{\tiny 0.5} &  80.6{\tiny$\pm$} &{\tiny 0.3} & 332.8{\tiny$\pm$} &{\tiny 1.0}  \\ 
3:~\resnet{18}{}{\iio}	 					& CO          & 84.8{\tiny$\pm$} &{\tiny 0.2} & 343.5{\tiny$\pm$} &{\tiny 2.7} &  73.1{\tiny$\pm$} &{\tiny 0.3} & 365.7{\tiny$\pm$} &{\tiny 2.8} &  78.6{\tiny$\pm$} &{\tiny 1.0} & 336.2{\tiny$\pm$} &{\tiny 3.5}  \\ 
4:~\resnet{18}{}{\hatghw}	 		    & CO          & 83.7{\tiny$\pm$} &{\tiny 0.9} & 354.4{\tiny$\pm$} &{\tiny 2.0} &  70.0{\tiny$\pm$} &{\tiny 1.7} & 380.7{\tiny$\pm$} &{\tiny 2.9} &  74.2{\tiny$\pm$} &{\tiny 3.6} & 358.6{\tiny$\pm$} &{\tiny 5.5}  \\   
\red{5:~\resnet{18}{}{\hatghw}}   & \red{CE}    & \red{75.5{\tiny$\pm$}} &\red{{\tiny 1.3}} & \red{391.0{\tiny$\pm$}} &\red{{\tiny 8.2}} &  \red{63.7{\tiny$\pm$}} &\red{{\tiny 1.6}} & \red{442.3{\tiny$\pm$}} &\red{{\tiny 9.7}} &  \red{64.0{\tiny$\pm$}} &\red{{\tiny 1.8}} & \red{427.5{\tiny$\pm$}} &\red{{\tiny 15.6}}  \\  
6:~\resnet{18}{}{\iio}	 					& CE          & 77.1{\tiny$\pm$} &{\tiny 0.9} & 375.0{\tiny$\pm$} &{\tiny 8.5} &  67.0{\tiny$\pm$} &{\tiny 1.0} & 429.0{\tiny$\pm$} &{\tiny 13.8} &  67.3{\tiny$\pm$} &{\tiny 2.1} & 417.8{\tiny$\pm$} &{\tiny 11.7}  \\  
7:~\resnet{18}{}{\ioi}	          & CE          & 78.4{\tiny$\pm$} &{\tiny 0.8} & 354.6{\tiny$\pm$} &{\tiny 10.5} &  67.8{\tiny$\pm$} &{\tiny 1.3} & 402.8{\tiny$\pm$} &{\tiny 12.2} &  66.7{\tiny$\pm$} &{\tiny 1.5} & 367.2{\tiny$\pm$} &{\tiny 14.8}  \\ 
8:~\resnet{18}{}{\oii}            & CE          & 77.0{\tiny$\pm$} &{\tiny 0.9} & 279.6{\tiny$\pm$} &{\tiny 5.6} &  65.4{\tiny$\pm$} &{\tiny 0.5} & 320.6{\tiny$\pm$} &{\tiny 6.8} &  68.6{\tiny$\pm$} &{\tiny 1.8} & 300.9{\tiny$\pm$} &{\tiny 11.4}  \\
9:~\resnet{18}{}{\ghw}	          & CE          & 80.4{\tiny$\pm$} &{\tiny 0.5} & 308.3{\tiny$\pm$} &{\tiny 4.0} &  69.9{\tiny$\pm$} &{\tiny 1.4} & 345.4{\tiny$\pm$} &{\tiny 5.2} &  71.1{\tiny$\pm$} &{\tiny 1.8} & 308.4{\tiny$\pm$} &{\tiny 4.1}  \\ 
\hline  
\end{tabular}
\end{center}

%% file: fig/train_curve.tex
\input{fig/data}
\begin{tikzpicture}
\begin{axis}[%
	width=0.55\linewidth,
	height=0.45\linewidth,
	xlabel={\small Epoch},
	ylabel={\small Training loss},
  axis y line*=left,
	xmin = -1,
	xmax = 21,
	grid=both,
	xlabel shift = -1ex,    
  error bars/y explicit = true,
]
	\addplot[color=brown,solid, mark=*,  mark size=0.7, line width=1.0, error bars/.cd, error mark=-,y dir=both] table[x=epoch, y expr={\thisrow{loss}}, y error expr={5*\thisrow{std}}]\traincurvelossstd;\label{otherplot}
\end{axis}
\begin{axis}[%
	width=0.55\linewidth,
	height=0.45\linewidth,
	xlabel={\small Epoch},
	ylabel={\small mAP on validation},
  axis y line*=right,
  hide x axis,
	legend cell align={left},
	legend pos=south east,
	legend style={at={(1.9,0)},cells={anchor=east}, font =\scriptsize, fill opacity=0.99, row sep=-2.5pt},
	xmin = -1,
	xmax = 21,
	grid=both,
	xlabel shift = -1ex,    
  error bars/y explicit = true,
]
  \addlegendimage{/pgfplots/refstyle=otherplot}\addlegendentry{Loss}
	\addplot[color=blue,     solid, mark=*,  mark size=0.7, line width=1.0, error bars/.cd, error mark=-,y dir=both] table[x=epoch, y expr={\thisrow{mAP}}, y error expr={5*\thisrow{std}}] \traincurvelocalstd;\leg{Local descriptors};
	\addplot[color=red,     solid, mark=*,  mark size=0.7, line width=1.0, error bars/.cd, error mark=-,y dir=both] table[x=epoch, y expr={\thisrow{mAP}}, y error expr={5*\thisrow{std}}] \traincurveglobalstd;\leg{Global descriptor};
	%
\end{axis}
\end{tikzpicture}

%% file: tab/rop.tex
%
\newcommand{\xdagger}{^{\dagger}}
\newcommand{\gemap}{GeM$_{\mbox{\tiny A\!P}}$\cite{rar+19}}
\newcommand{\gemapno}{GeM$_{\mbox{\tiny A\!P}}$}
\def\arraystretch{1.0}
\scriptsize
\begin{tabular}	{l@{\msp}r@{\msp}r@{\msp}c@{\msp}c@{\msp}c@{\msp}c@{\msp}c@{\msp}c@{\msp}c@{\msp}c@{\msp}c@{\msp}c}
\hline
\multirow{2}{*}{Method}        &    \multirow{2}{*}{FCN} & Mem & \multicolumn{2}{c}{Mean}  &   \multicolumn{2}{c}{\ro} & \multicolumn{2}{c}{\ro\hspace{-3pt}+\rdis} & \multicolumn{2}{c}{\rpa} & \multicolumn{2}{c}{\rp\hspace{-3pt}+\rdis}     \\ \cline{4-13}
															 &    												    &  (GB) & all & \rdis &   med  &   hard         &      med  &   hard          &    med  &   hard      &           med  &   hard         \\ \hline \hline
															\multicolumn{13}{c}{Global descriptors \& Euclidean distance search} \\ \hline \hline
R-MAC~\cite{gar+17}                                             & \resnet{101}{}{}     & 7.6 & 45.8 & 33.6 & 60.9 & 32.4 & 39.3 & 12.5 & 78.9 & 59.4 & 54.8 & 28.0 \\
GeM~\cite{rtc19}                                                & \resnet{101}{}{}     & 7.6 & 47.4 & 35.5 & 64.7 & 38.5 & 45.2 & 19.9 & 77.2 & 56.3 & 52.3 & 24.7 \\
GeM~\cite{rtc19}$\xdagger$                                      & \resnet{101}{}{}     & 7.6 & 47.3 & 35.2 & 65.4 & 40.1 & 45.1 & 22.7 & 76.7 & 55.2 & 50.8 & 22.4 \\       
GeM$_w$$\xdagger$                                              & \resnet{101}{}{}     & 7.6 & 49.0 & 37.2 & 67.8 & 41.7 & 47.7 & 23.3 & 77.6 & 56.3 & 52.9 & 25.0 \\
\gemap                                      & \resnet{101}{}{}     & 7.6 & 49.9 & 37.1 & 67.5 & 42.8 & 47.5 & 23.2 & 80.1 & 60.5 & 52.5 & 25.1 \\
\gemap$\xdagger$                                  & \resnet{101}{}{}     & 7.6 & 49.7 & 36.7 & 67.1 & 42.3 & 47.8 & 22.5 & 80.3 & 60.9 & 51.9 & 24.6 \\             
\gemap\,{\tiny {PQ1}}                    & \resnet{101}{}{}      & 1.9 & 49.6 & 36.7 & 67.1 & 42.2 & 47.7 & 22.5 & 80.3 & 60.8 & 51.9 & 24.6 \\
\gemap\,{\tiny {PQ8}}                    & \resnet{101}{}{}      & 0.2 & 48.1 & 35.5 & 65.1 & 40.4 & 46.1 & 21.6 & 78.7 & 58.7 & 50.7 & 23.5 \\
\gemap$\xdagger$\,{\tiny $\triangleright1024$}                               & \resnet{101}{}{}     & 3.8 & 49.0 & 35.8 & 66.6 & 41.6 & 46.7 & 21.7 & 80.0 &   60.3 & 51.0 & 23.8 \\             
\gemap$\xdagger$\,{\tiny $\triangleright512$}                                & \resnet{101}{}{}     & 1.9 & 47.5 & 33.7 & 65.9 & 40.5 & 43.9 & 19.7 & 79.5 & 59.4 & 48.9 & 22.3 \\             
\gemapno + GeM$_w$$\xdagger$            & \resnet{101}{}{}      & 15.3 & 53.4 & 41.4 & 70.5 & 45.7 & 52.6 & 27.1 & \textbf{81.9} & \textbf{63.4} & 57.0 & 29.1 \\
\hline \hline
															\multicolumn{13}{c}{Local descriptors \& ASMK} \\ \hline \hline
DELF~\cite{nas+17}$\xdagger$                                    & \resnet{50}{-}{}     & 9.2 & 53.0 & 43.1 & 69.0 & 44.0 & 54.1 & 31.1 & 79.5 & 58.9 & 59.3 & 28.1 \\   
DELF~\cite{nas+17} $\ddagger$                                   & \resnet{50}{-}{}     & 9.2 & 52.5 & 42.7 & 69.2 & 44.3 & 54.3 & 31.3 & 78.7 & 57.4 & 58.2 & 26.9 \\   
DELF~\cite{nas+17}\cite{rit+18}         & \resnet{50}{-}{}     & 9.2 & 51.5 & 42.2 & 67.8 & 43.1 & 53.8 & 31.2 & 76.9 & 55.4 & 57.3 & 26.4 \\   
\resnet{18}{}{\ghw}, $n=1000$                                   & \resnet{18}{}{}      & 4.6 & 54.8 & 43.5 & 75.1 & 51.7 & 55.7 & 32.0 & 79.4 & 58.3 & 57.4 & 28.9 \\  
\resnet{50}{-}{\ghw}, $n=1000$                                  & \resnet{50}{-}{}     & 7.9 & 58.0 & 47.4 & 78.3 & 55.8 & 63.6 & 36.8 & 80.1 & 60.1 & 58.4 & 30.7 \\  
\resnet{50}{-}{\ghw}, $n=1200$                                  & \resnet{50}{-}{}     & 9.2 & 58.8 & 48.4 & 78.8 & 56.7 & 64.5 & 37.7 & 80.6 & 61.0 & 59.6 & 31.7 \\  
\resnet{50}{-}{\ghw}, $n=1400$                                  & \resnet{50}{-}{}     & 10.6 & 59.3 & 49.0 & 79.1 & 56.8 & 64.9 & 38.2 & 81.0 & 61.5 & 60.4 & 32.6 \\  
\resnet{50}{-}{\ghw}, $n=2000$                                  & \resnet{50}{-}{}     & 14.3 & \textbf{60.1} & \textbf{50.1} & \textbf{79.4} & \textbf{56.9} & \textbf{65.8} & \textbf{38.9} & 81.6 & 62.4 & \textbf{61.8} & \textbf{33.7} \\  
 \hline
\end{tabular}
%


%% file: tab/gl.tex
%
\def\arraystretch{1.0}
\begin{tabular}{l@{\lsp}r@{\lsp}r@{\lsp}r@{\lsp}r@{\lsp}r@{\lsp}r}
\hline
Method                                  &   FCN    &      Training set            &     Memory (GB)  &     CLS1   &     CLS2  	& CLS3   \\ \hline \hline
GEM~\cite{rtc19}                        &   \resnet{101}{}{}          &      SfM-120k						    &					31.5     &    1.9     &    18.0     &   24.1     \\
GEM$_w$                                 &   \resnet{101}{}{}          &      SfM-120k  					    &					31.5  	 &    3.7     &    23.4     &   28.7     \\    
GEM-AP~\cite{rar+19}                     &   \resnet{101}{}{}          &      SfM-120k						    &					31.5  	 &    2.8     &    14.8     &   20.7     \\ 
DELF~\cite{nas+17}                      &   \resnet{50}{-}{}          &   Landmarks~\cite{bsc+14}	  &  					24.1		 &    2.1     &    11.9     &   21.9     \\
\resnet{18}{}{\ghw}                &   \resnet{18}{}{}           &      SfM-120k            	  &					17.5  	 &    8.5     &    20.0     &   27.0     \\    
\resnet{50}{-}{\ghw}               &   \resnet{50}{-}{}          &      SfM-120k               &					29.3   	 &    18.5    &    33.1     &   36.5     \\    
\hline
\end{tabular}
%


%% file: paper.bbl
\begin{thebibliography}{10}
\providecommand{\url}[1]{\texttt{#1}}
\providecommand{\urlprefix}{URL }
\providecommand{\doi}[1]{https://doi.org/#1}

\bibitem{az13}
Arandjelovi\'c, R., Zisserman, A.: All about {VLAD}. In: CVPR (2013)

\bibitem{az14}
Arandjelovi\'c, R., Zisserman, A.: {DisLocation}: {Scalable} descriptor
  distinctiveness for location recognition. In: ACCV (2014)

\bibitem{agt+15}
Arandjelovi\'{c}, R., Gronat, P., Torii, A., Pajdla, T., Sivic, J.: {NetVLAD}:
  {CNN} architecture for weakly supervised place recognition. In: CVPR (2016)

\bibitem{bl15}
Babenko, A., Lempitsky, V.: Aggregating deep convolutional features for image
  retrieval. In: ICCV (2015)

\bibitem{bsc+14}
Babenko, A., Slesarev, A., Chigorin, A., Lempitsky, V.: Neural codes for image
  retrieval. In: ECCV (2014)

\bibitem{blv+17}
Balntas, V., Lenc, K., Vedaldi, A., Mikolajczyk, K.: Hpatches: A benchmark and
  evaluation of handcrafted and learned local descriptors. In: CVPR (2017)

\bibitem{brp+19}
Barroso~Laguna, A., Riba, E., Ponsa, D., Mikolajczyk, K.: Key. net: Keypoint
  detection by handcrafted and learned cnn filters. In: ICCV (2019)

\bibitem{bet+08}
Bay, H., Ess, A., Tuytelaars, T., Gool, L.V.: {SURF}: Speeded up robust
  features. Computer Vision and Image Understanding  \textbf{110}(3),  346--359
  (May 2008)

\bibitem{bgp19}
Benbihi, A., Geist, M., Pradalier, C.: Elf: Embedded localisation of features
  in pre-trained cnn. In: CVPR (2019)

\bibitem{bgr+20}
Bhowmik, A., Gumhold, S., Rother, C., Brachmann, E.: Reinforced feature points:
  Optimizing feature detection and description for a high-level task. In: CVPR
  (2020)

\bibitem{cas20}
Cao, B., Araujo, A., Sim, J.: Unifying deep local and global features for
  efficient image search. In: arxiv (2020)

\bibitem{dmr18}
DeTone, D., Malisiewicz, T., Rabinovich, A.: Superpoint: Self-supervised
  interest point detection and description. In: CVPRW (2018)

\bibitem{drp+19}
Dusmanu, M., Rocco, I., Pajdla, T., Pollefeys, M., Sivic, J., Torii, A.,
  Sattler, T.: D2-net: A trainable cnn for joint detection and description of
  local features. In: CVPR (2019)

\bibitem{gar+17}
Gordo, A., Almazan, J., Revaud, J., Larlus, D.: End-to-end learning of deep
  visual representations for image retrieval. IJCV  (2017)

\bibitem{glj19}
Gu, Y., Li, C., Jiang, Y.G.: Towards optimal cnn descriptors for large-scale
  image retrieval. In: ACM Multimedia (2019)

\bibitem{hb16}
Husain, S., Bober, M.: Improving large-scale image retrieval through robust
  aggregation of local descriptors. PAMI  \textbf{39}(9),  1783--1796 (Jan
  2016)

\bibitem{itg+15}
Iscen, A., Tolias, G., Gosselin, P.H., J{\'e}gou, H.: A comparison of dense
  region detectors for image search and fine-grained classification. {IEEE}
  Transactions on Image Processing  \textbf{24}(8),  2369--2381 (2015)

\bibitem{jds09}
J\'egou, H., Douze, M., Schmid, C.: On the burstiness of visual elements. In:
  CVPR (Jun 2009)

\bibitem{jc12}
J\'egou, H., Chum, O.: Negative evidences and co-occurences in image retrieval:
  The benefit of {PCA} and whitening. In: ECCV (Oct 2012)

\bibitem{jds10}
J\'egou, H., Douze, M., Schmid, C.: Improving bag-of-features for large scale
  image search. IJCV  \textbf{87}(3),  316--336 (Feb 2010)

\bibitem{jds11}
J\'egou, H., Douze, M., Schmid, C.: Product quantization for nearest neighbor
  search. PAMI  \textbf{33}(1),  117--128 (Jan 2011)

\bibitem{jpd+12}
J\'egou, H., Perronnin, F., Douze, M., S\'anchez, J., P\'erez, P., Schmid, C.:
  Aggregating local descriptors into compact codes. In: PAMI (Sep 2012)

\bibitem{kmo16}
Kalantidis, Y., Mellina, C., Osindero, S.: Cross-dimensional weighting for
  aggregated deep convolutional features. In: ECCVW (2016)

\bibitem{kdf17}
Kim, H.J., Dunn, E., Frahm, J.M.: Learned contextual feature reweighting for
  image geo-localization. In: CVPR (2017)

\bibitem{mcm+04}
Matas, J., Chum, O., Urban, M., Pajdla, T.: Robust wide-baseline stereo from
  maximally stable extremal regions. Image and Vision Computing
  \textbf{22}(10),  761--767 (2004)

\bibitem{ms05}
Mikolajczyk, K., Schmid, C.: A performance evaluation of local descriptors.
  PAMI  \textbf{27}(10),  1615--1630 (2005)

\bibitem{mts+05}
Mikolajczyk, K., Tuytelaars, T., Schmid, C., Zisserman, A., Matas, J.,
  Schaffalitzky, F., Kadir, T., Gool, L.V.: A comparison of affine region
  detectors. IJCV  \textbf{65}(1/2),  43--72 (2005)

\bibitem{mmo+16}
Mohedano, E., McGuinness, K., O'Connor, N.E., Salvador, A., Marques, F., Giro-i
  Nieto, X.: Bags of local convolutional features for scalable instance search.
  In: ICMR (2016)

\bibitem{nas+17}
Noh, H., Araujo, A., Sim, J., Weyand, T., Han, B.: Large-scale image retrieval
  with attentive deep local features. In: ICCV (2017)

\bibitem{pls+10}
Perronnin, F., Liu, Y., Sanchez, J., Poirier, H.: Large-scale image retrieval
  with compressed {Fisher} vectors. In: CVPR (2010)

\bibitem{plr09}
Perronnin, F., Liu, Y., Renders, J.M.: A family of contextual measures of
  similarity between distributions with application to image retrieval. In:
  CVPR. pp. 2358--2365 (2009)

\bibitem{pci+07}
Philbin, J., Chum, O., Isard, M., Sivic, J., Zisserman, A.: Object retrieval
  with large vocabularies and fast spatial matching. In: CVPR (2007)

\bibitem{pci+08}
Philbin, J., Chum, O., Isard, M., Sivic, J., Zisserman, A.: Lost in
  quantization: Improving particular object retrieval in large scale image
  databases. In: CVPR (Jun 2008)

\bibitem{rit+18}
Radenovi{\'c}, F., Iscen, A., Tolias, G., Avrithis, Y., Chum, O.: Revisiting
  oxford and paris: Large-scale image retrieval benchmarking. In: CVPR (2018)

\bibitem{rtc19}
Radenovi{\'c}, F., Tolias, G., Chum, O.: Fine-tuning {CNN} image retrieval with
  no human annotation. PAMI  \textbf{41} (Jul 2019)

\bibitem{rsa+14}
Razavian, A.S., Sullivan, J., Carlsson, S., Maki, A.: Visual instance retrieval
  with deep convolutional networks. ITE Trans. on Media Technology and
  Applications  (2016)

\bibitem{rar+19}
Revaud, J., Almaz{\'{a}}n, J., de~Rezende, R.S., de~Souza, C.R.: Learning with
  average precision: Training image retrieval with a listwise loss. In: ICCV
  (2019)

\bibitem{rwd+19}
Revaud, J., Weinzaepfel, P., De~Souza, C., Pion, N., Csurka, G., Cabon, Y.,
  Humenberger, M.: R2d2: Repeatable and reliable detector and descriptor. In:
  NeurIPS (2019)

\bibitem{rds+15}
Russakovsky, O., Deng, J., Su, H., Krause, J., Satheesh, S., Ma, S., Huang, Z.,
  Karpathy, A., Khosla, A., Bernstein, M., et~al.: Imagenet large scale visual
  recognition challenge. IJCV  (2015)

\bibitem{src+15}
Sch\"{o}nberger, J.L., Radenovi\'{c}, F., Chum, O., Frahm, J.M.: From single
  image query to detailed {3D} reconstruction. In: CVPR (2015)

\bibitem{sac19}
Sim{\'e}oni, O., Avrithis, Y., Chum, O.: Local features and visual words emerge
  in activations. In: CVPR (2019)

\bibitem{sz03}
Sivic, J., Zisserman, A.: {Video Google}: {A} text retrieval approach to object
  matching in videos. In: ICCV (2003)

\bibitem{taz+19}
Teichmann, M., Araujo, A., Zhu, M., Sim, J.: Detect-to-retrieve: Efficient
  regional aggregation for image search. In: CVPR (2019)

\bibitem{taj16}
Tolias, G., Avrithis, Y., J\'egou, H.: Image search with selective match
  kernels: aggregation across single and multiple images. IJCV  (2015)

\bibitem{tsj16}
Tolias, G., Sicre, R., J{\'e}gou, H.: Particular object retrieval with integral
  max-pooling of {CNN} activations. In: ICLR (2016)

\bibitem{vjh17}
Vo, N., Jacobs, N., Hays, J.: Revisiting im2gps in the deep learning era. In:
  CVPR (2017)

\bibitem{wzh+20}
Wang, Q., Zhou, X., Hariharan, B., Snavely, N.: Learning feature descriptors
  using camera pose supervision. In: arXiv (2020)

\bibitem{wac+20}
Weyand, T., Araujo, A., Cao, B., Sim, J.: Google landmarks dataset v2-a
  large-scale benchmark for instance-level recognition and retrieval. In: CVPR
  (2020)

\bibitem{ynh+20}
Yang, T., Nguyen, D., Heijnen, H., Balntas, V.: Ur2kid: Unifying retrieval,
  keypoint detection, and keypoint description without local correspondence
  supervision. In: arxiv (2020)

\bibitem{yyd15}
Yue-Hei~Ng, J., Yang, F., Davis, L.S.: Exploiting local features from deep
  networks for image retrieval. In: CVPR (2015)

\bibitem{zjs13}
Zhu, C.Z., J{\'e}gou, H., ichi Satoh, S.: Query-adaptive asymmetrical
  dissimilarities for visual object retrieval. In: ICCV (2013)

\end{thebibliography}
